\pdfoutput=1

\documentclass[11pt]{article}

\usepackage{acl}

\usepackage{times}
\usepackage{latexsym}

\usepackage[T1]{fontenc}

\usepackage[utf8]{inputenc}

\usepackage{microtype}

\usepackage{mathpartir}

\usepackage{times}
\usepackage{latexsym}

\usepackage[T1]{fontenc}

\usepackage[utf8]{inputenc}
\usepackage{microtype}

\usepackage{inconsolata}
\usepackage{amsmath,amsthm,verbatim,amssymb,amsfonts,amscd, graphicx}

\theoremstyle{plain}

\theoremstyle{definition}

\usepackage{amsmath}
\usepackage{proof}
\usepackage{blkarray}

\usepackage{url}

\newcommand{\R}{\mathcal{R}}

\usepackage{listings}
\definecolor{sclgreen}{rgb}{0,0.46,0}
\definecolor{scllightgrey}{rgb}{0.94,0.94,0.94}%
\definecolor{sclgreyblue}{rgb}{0.3,0.4,0.6}%
\definecolor{sclcyan}{rgb}{0.1,0.4,0.6}%
\definecolor{sclpurple}{rgb}{0.71,0,0.85}%
\definecolor{sclyellow}{rgb}{0.9,0.6,0.05}%
\definecolor{sclorange}{rgb}{1,0.36,0.03}%
\definecolor{sclred}{rgb}{0.6,0.2,0.0}%

\lstdefinelanguage{scallop}{
    keywords={import,type,const,rel,usize,where,String,i8,i32,i64,usize,u8,u16,u32,u64},keywordstyle=\color{blue},%
    morekeywords=[2]{and,or,not,==,+,-,*,/},keywordstyle=[2]\color{sclpurple},
    morekeywords=[3]{count, exists, sum, prod, forall, as},
    keywordstyle=[3]\color{sclorange},
    morecomment=[s]{/*}{*/},
    commentstyle=\color{sclgreen},%
    morecomment=[l]{//},%
    morestring=[b]",stringstyle=\color{sclorange}
}

\usepackage{amsmath,amsthm,verbatim,amssymb,amsfonts,amscd}

\usepackage{microtype} %
\usepackage[bottom]{footmisc}
\usepackage{caption}
\usepackage{subcaption}

\usepackage{helvet}  %
\usepackage{courier}  %
\usepackage{url}  %
\usepackage{graphicx}  %
\usepackage{multirow}
\usepackage{amsthm}
\usepackage{color}
\usepackage{MnSymbol}
\usepackage{makecell}
\usepackage{arydshln}
\usepackage{amsmath}
\usepackage{xcolor}
\usepackage{natbib}
\usepackage{adjustbox}
\usepackage{booktabs}
\usepackage{textcomp}
\usepackage{wrapfig}
\usepackage{algorithm}
\usepackage{algorithmic}
\usepackage{listings}
\usepackage{csquotes}
\usepackage{tcolorbox}
\newcommand{\abbr}[0]{DSR-LM} %

\usepackage{enumitem}
\usepackage{tikz}
\usepackage{pgfplots}

\definecolor{mygreen}{HTML}{D5E8D4}
\definecolor{myred}{HTML}{F8CECC}
\definecolor{myorange}{HTML}{FFE6CC}
\definecolor{myblue}{HTML}{BAE8FC}
\definecolor{mypurple}{HTML}{E1D5E7}
\definecolor{myyellow}{HTML}{FFF2CC}
\definecolor{mygrey}{HTML}{e8e8e8}

\definecolor{mydeepgreen}{HTML}{82B366}
\definecolor{mydeepred}{HTML}{B85450}
\definecolor{mydeeporange}{HTML}{f29202}
\definecolor{mydeepblue}{HTML}{6C8EBF}
\definecolor{mydeeppurple}{HTML}{9673A6}
\definecolor{mydeepyellow}{HTML}{D6B656}
\definecolor{mydeepgrey}{HTML}{b8b8b8}

\newcommand{\figref}[1]{Figure~\ref{#1}}
\newcommand{\tabref}[1]{Table~\ref{#1}}

\title{Improved Logical Reasoning of Language Models via Differentiable Symbolic Programming}

\author{Hanlin Zhang$^{1,}$\thanks{~~Equal contribution} \ 
  Jiani Huang$^{2,*}$ 
  Ziyang Li$^{2}$
Mayur Naik$^{2}$
  Eric Xing$^{1,3,4}$ \\
  $^1$Carnegie Mellon University, 
  $^2$University of Pennsylvania,  
  \\
  $^3$Mohamed Bin Zayed University of Artificial Intelligence, $^4$Petuum Inc. 
  \\
}

\begin{document}
\setlength{\abovedisplayskip}{5pt}
\setlength{\belowdisplayskip}{5pt}
\maketitle

\begin{abstract}
Pre-trained large language models (LMs) struggle to perform logical reasoning reliably despite advances in scale and compositionality. In this work, we tackle this challenge through the lens of symbolic programming. We propose DSR-LM, a \textbf{\underline{D}}ifferentiable \textbf{\underline{S}}ymbolic \textbf{\underline{R}}easoning framework where pre-trained \textbf{\underline{LM}}s govern the perception of factual knowledge, and a symbolic module
performs deductive reasoning. In contrast to works that rely on hand-crafted logic rules, our differentiable symbolic reasoning framework efficiently learns weighted rules and applies semantic loss to further improve LMs. 
DSR-LM is scalable, interpretable, and allows easy integration of prior knowledge, thereby supporting extensive symbolic programming to robustly derive a logical conclusion.
The results of our experiments suggest that DSR-LM improves the logical reasoning abilities of pre-trained language models, resulting in a significant increase in accuracy of over 20\% on deductive reasoning benchmarks. 
Furthermore, DSR-LM outperforms a variety of competitive baselines when faced with systematic changes in sequence length.\footnote{Code available at \href{https://github.com/moqingyan/dsr-lm}{https://github.com/moqingyan/dsr-lm}}
\end{abstract}

\vspace{-2mm}
\section{Introduction}

Complex applications in natural language processing involve dealing with two separate challenges.
On one hand, there is the richness, nuances, and extensive vocabulary of natural language.
On the other hand, one needs logical connectives, long reasoning chains, and domain-specific knowledge to draw logical conclusions.
The systems handling these two challenges are complementary to each other and are likened to psychologist Daniel Kahneman's human ``system~1'' and ``system~2'' \cite{kahneman2011thinking}:
while the former makes fast and intuitive decisions, akin to neural networks, the latter thinks more rigorously and methodically.
Considering LMs as ``system 1'' and symbolic reasoners as ``system 2'', we summarize their respective advantages in Table~\ref{tab:neural_symbolic_advantage}.

Although pre-trained LMs have demonstrated remarkable predictive performance, making them an effective ``system 1'', they fall short when asked to perform consistent logical reasoning~\citep{kassner2020pretrained, helwe2021reasoning, creswell2022selection}, which usually requires ``system 2''.
In part, this is because LMs largely lack capabilities of systematic generalization~\citep{elazar2021measuring, hase2021language, valmeekam2022still}.

\begin{table}[t]
  \centering
  \footnotesize
  \begin{tabular}{p{0.45\linewidth}|p{0.45\linewidth}}
    \toprule \hline
    \multicolumn{1}{c|}{\textbf{Language Model}} & \multicolumn{1}{c}{\textbf{Symbolic Reasoner}} \\
    \hline
    \begin{itemize}[leftmargin=5pt,noitemsep,topsep=0pt,parsep=0pt]
      \item Rapid reasoning
      \item Sub-symbolic knowledge
      \item Handling noise, ambiguities, and naturalness
      \item Process open domain text
      \item Can learn in-context
    \end{itemize}
    &
    \begin{itemize}[leftmargin=7pt,noitemsep,topsep=0pt,parsep=0pt]
      \item Multi-hop reasoning
      \item Compositionality
      \item Interpretability
      \item Data efficiency
      \item Can incorporate domain-specific knowledge
    \end{itemize}
\\
    \bottomrule
\end{tabular}
\captionof{table}{Respective advantages of \textbf{language models} and \textbf{symbolic reasoners}.}
\label{tab:neural_symbolic_advantage}
\vspace{-0.2in}
\end{table}

In this work, we seek to incorporate deductive logical reasoning with LMs.
Our approach has the same key objectives as neuro-symbolic programming~\citep{chaudhuri2021neurosymbolic}: compositionality, consistency, interpretability, and easy integration of prior knowledge.
We present \abbr, which tightly integrates a differentiable symbolic reasoning module with pre-trained LMs in an end-to-end fashion.
With \abbr, the underlying LMs govern the perception of natural language and are fine-tuned to extract relational triplets with only weak supervision.
To overcome a common limitation of symbolic reasoning systems, the reliance on human-crafted logic rules~\citep{huang2021scallop, nye2021improving}, we adapt \abbr~to induce and fine-tune rules automatically.
Further, \abbr~ allows incorporation of semantic loss obtained by logical integrity constraints given as prior knowledge, which substantially helps the robustness.

We conduct extensive experiments showing that \abbr~can consistently improve the logical reasoning capability upon pre-trained LMs.
Even if \abbr~uses a RoBERTa backbone with much less parameters and does not explicitly take triplets as supervision, it can still outperform various baselines by large margins.
Moreover, we show that \abbr~can induce logic rules that are amenable to human understanding to explain decisions given only higher-order predicates.
As generalization over long-range dependencies is a significant weakness of transformer-based language models \citep{lake2018generalization, tay2020long}, we highlight that in systematic, long-context scenarios, where most pre-trained or neural approaches fail to generalize compositionally, \abbr~can still achieve considerable performance gains.

\begin{figure*}[ht]
    \centering
    \includegraphics[width=\textwidth]{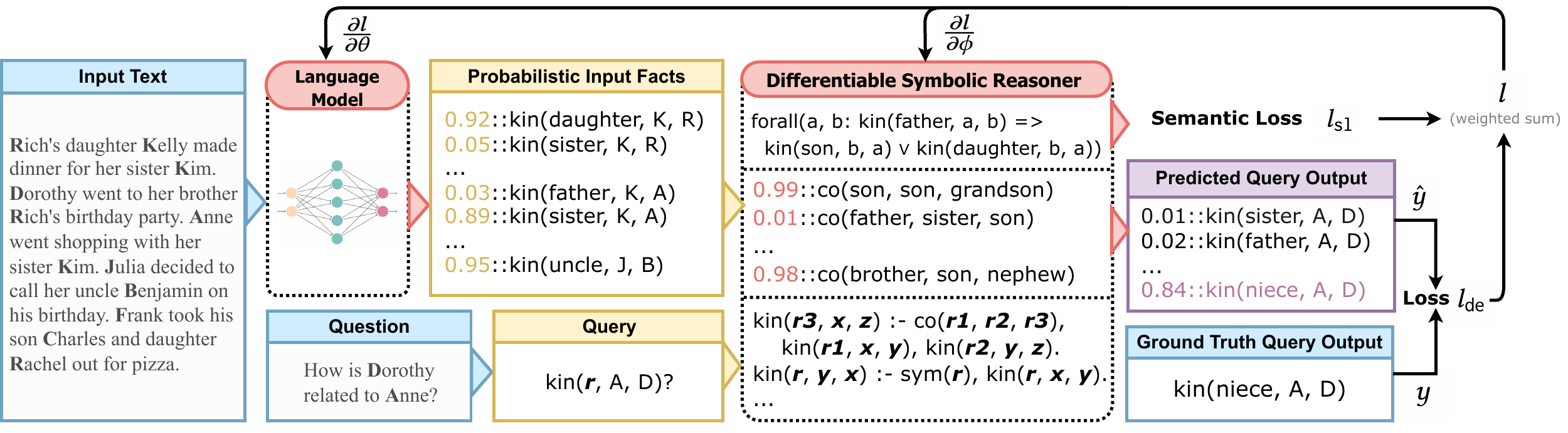}
    \caption{
        Overview of \abbr~with a motivating example where ``Anne is the niece of Dorothy'' should be logically inferred from the context.
        We abbreviate the names with their first initials in the relational symbols.
    }
    \label{fig:moti}
    \vspace{-0.1in}
\end{figure*}

\vspace{-2mm}
\section{Related Work}
\label{sec:related}
\textbf{Logical reasoning with LMs.}
Pre-trained LMs have been shown to struggle with logical reasoning over factual knowledge \citep{kassner2020pretrained, helwe2021reasoning, talmor2020olmpics}.
There is encouraging recent progress in using transformers for reasoning tasks \citep{zhou2020towards, clark2021transformers, wei2022chain, schowdhery2022palm, zelikman2022star} but these approaches usually require a significant amount of computation for re-training or human annotations on reasoning provenance~\citep{camburu2018snli, zhou2020towards, nye2021improving, wei2022chain}.
Moreover, their entangled nature with natural language makes it fundamentally hard to achieve robust inference over factual knowledge~\citep{greff2020binding, saparov2022language, zhang2022impact}.

There are other obvious remedies for LMs' poor reasoning capability.
Ensuring that the training corpus contains a sufficient amount of exemplary episodes of sound reasoning reduces the dependency on normative biases and annotation artifacts~\citep{talmor2020leap, betz2020critical, hase2021language}.
Heuristics like data augmentation are also shown to be effective~\citep{talmor2020leap}.
But the above works require significant efforts for crowdsourcing and auditing training data. Our method handily encodes a few prototypes/templates of logic rules and is thus more efficient in terms of human effort.
Moreover, our goal is fundamentally different from theirs in investigating the tight integration of neural and symbolic models in an end-to-end manner.

\textbf{Neuro-symbolic reasoning.} 
Neuro-symbolic approaches are proposed to integrate the perception of deep neural components and the reasoning of symbolic components.
Representative works can be briefly categorized into regularization~\citep{xu2018semantic}, program synthesis~\citep{mao2018neuro}, and proof-guided probabilistic programming~\citep{evans2018learning, rocktaschel2017end, manhaeve2018deepproblog, zhang2019efficient, huang2021scallop}.
To improve compositionality of LMs, previous works propose to parameterize grammatical rules \citep{kim2021sequence, shaw2021compositional} but show that those hybrid models are inefficient and usually underperform neural counterparts.
In contrast to the above works, \abbr~focuses on improving LMs' reasoning over logical propositions with tight integration of their pre-trained knowledge in a scalable and automated way.

\vspace{-2mm}
\definecolor{scllightgrey}{rgb}{0.94,0.94,0.94}%
\definecolor{sclgreyblue}{rgb}{0.2,0.3,0.5}%
\newcommand{\sclcode}[1]{{\tcbox[on line,boxsep=0pt,left=1pt,right=1pt,top=1pt,bottom=1pt,colframe=scllightgrey,colback=scllightgrey]{\color{sclgreyblue} \small \texttt{#1}}}}
\newcommand{\sclcodefn}[1]{{\tcbox[on line,boxsep=0pt,left=1pt,right=1pt,top=1pt,bottom=1pt,colframe=scllightgrey,colback=scllightgrey]{\color{sclgreyblue} \scriptsize \texttt{#1}}}}

\section{Methodology}

\subsection{Problem Formulation}

Each question answering (QA) example in the dataset is a triplet containing input text $x$, query $q$, and the answer $y$.
Figure \ref{fig:moti} shows an instance that we will use as our running example.
The input text $x$ is a natural language passage within which there will be a set of entities, possibly referenced by 3rd person pronouns.
The sentences hint at the relationships between entities.
For example, ``Dorothy went to her brother Rich's birthday party'' implies that Rich is Dorothy's brother and Dorothy is Rich's sister.
The query $q$ is a tuple of two entities, representing the people with whom we want to infer the relation.
The expected relation is stored in the answer $y$, which will be one of a confined set of possible relations $\R$, allowing us to treat the whole problem as an $|\R|$-way classification problem.
We focus only on the problems where the desired relation is not explicitly stated in the context but need to be deduced through a sequence of reasoning.

\subsection{Methodology Overview}

The design of DSR-LM concerns tightly integrating a perceptive model for relation extraction with a symbolic engine for logical reasoning.
While we apply LMs for low-level perception and relation extraction, we employ a symbolic reasoning module to consistently and logically reason about the extracted relations.
With a recent surge in neuro-symbolic methods, reasoning engines are made differentiable, allowing us to differentiate through the logical reasoning process.
In particular, we employ Scallop \cite{huang2021scallop} as our reasoning engine.
We propose two add-ons to the existing neuro-symbolic methodology.
First, some rules used for logical deduction are initialized using language models and further tuned by our end-to-end pipeline, alleviating human efforts.
Secondly, we employ integrity constraints on the extracted relation graphs and the logical rules, to improve the logical consistency of LMs and the learned rules.

Based on this design, we formalize our method as follows.
We adopt pretrained LMs to build relation extractors, denoted $\mathcal{M}_\theta$, which take in the natural language input $x$ and return a set of probabilistic relational symbols $\mathbf{r}$.
Next, we employ a differentiable deductive reasoning program, $\mathcal{P}_\phi$, where $\phi$ represents the weights of the learned logic rules.
It takes as input the probabilistic relational symbols and the query $q$ and returns a distribution over $\R$ as the output $\hat{y}$.
Overall, the deductive model is written as
\begin{align}
\hat{y} = \mathcal{P}_\phi(\mathcal{M}_\theta(x), q).
\end{align}
Additionally, we have the semantic loss (\texttt{sl}) derived by another symbolic program $\mathcal{P}_{\texttt{sl}}$ computing the probability of violating the integrity constraints:
\begin{align}
l_{\texttt{sl}} = \mathcal{P}_{\texttt{sl}}(\mathcal{M}_\theta(x), \phi)
\end{align}
Combined, we aim to minimize the objective $J$ over training set $\mathcal{D}$ with loss function $\mathcal{L}$:
\begin{equation}
\begin{aligned}[b]
J(\theta, \phi) &= \frac{1}{|\mathcal{D}|} \sum_{(x, q, y) \in \mathcal{D}} w_1 \mathcal{L}(\mathcal{P}_\phi(\mathcal{M}_\theta(x), q), y) \\
&+ w_2 \mathcal{P}_{\texttt{sl}}(\mathcal{M}_\theta(x), \phi),
\end{aligned}
\end{equation}
where $w_1$ and $w_2$ are tunable hyper-parameters to balance the deduction loss and semantic loss.

\subsection{Relation Extraction}

Since pre-trained LMs have strong pattern recognition capabilities for tasks like Named-Entity-Recognition (NER) and Relation Extraction (RE) \citep{tenney2019bert, soares2019matching}, we adopt them as our neural components in \abbr.
To ensure that LMs take in strings of similar length, we divide the whole context into multiple windows.
The goal is to extract the relations between every pair of entities in each windowed context.
Concretely, our relation extractor $\mathcal{M}_\theta$ comprises three components:
1) a Named-Entity Recognizer (NER) to obtain the entities in the input text,
2) a pre-trained language model, to be fine-tuned, that converts windowed text into embeddings, and
3) a classifier that takes in the embedding of entities and predicts the relationship between them.
The set of parameters $\theta$ contains the parameters of both the LM and the classifier.

We assume the relations to be classified come from a finite set of relations $\R$.
For example in CLUTRR \citep{sinha2019clutrr}, we have 20 kinship relations including mother, son, uncle, father-in-law, etc.
In practice, we perform $(|\R| + 1)$-way classification over each pair of entities, where the extra class stands for ``n/a''.
The windowed contexts are split based on simple heuristics of ``contiguous one to three sentences that contain at least two entities'', to account for coreference resolution.
The windowed contexts can be overlapping and we allow the reasoning module to deal with noisy and redundant data.
Overall, assuming that there are $m$ windows in the context $x$, we extract $m n (n - 1) (|\R| + 1)$ probabilistic relational symbols.
Each symbol is denoted as an atom of the form $p(s, o)$, where $p \in \R \cup \{\text{n/a}\}$ is the relational predicate, and $s$, $o$ are the two entities connected by the predicate.
We denote the probability of such symbol extracted by the LM and relational classifier as $\Pr(p(s, o) ~|~ \theta)$.
All these probabilities combined form the output vector $\mathbf{r} = \mathcal{M}_\theta(x) \in \mathbb{R}^{m n(n - 1) (|\R| + 1)}$.

\subsection{Differentiable Symbolic Inference}

The symbolic inference modules $\mathcal{P}_\phi$ and $\mathcal{P}_{\texttt{sl}}$ are responsible for processing the extracted relations to deduce 1) an expected output relation in $\R$, and 2) a semantic loss encoding the probability of constraint violation.
There are two main objectives for these modules.
First, they need to logically reason about the output relation and the semantic loss based on the extracted relational symbols $\mathbf{r}$, the query $q$, and the rule weights $\phi$.
Second, they need to compute the gradients of $\hat{y}$ and $l_\texttt{sl}$ with respect to $\theta$ and $\phi$, namely $\frac{\partial \hat{y}}{\partial \theta}$, $\frac{\partial \hat{y}}{\partial \phi}$, $\frac{\partial l_\texttt{sl}}{\partial \phi}$, and $\frac{\partial l_\texttt{sl}}{\partial \theta}$, in order for the fine-tuning and rule learning to happen.

\paragraph{Logical deduction.}

Logic rules can be applied to known facts to deduce new ones.
For example, below is a horn clause, which reads
``if $b$ is $a$'s brother and $c$ is $b$'s daughter, then $c$ is $a$'s niece'':
\begin{align*}
\small
\text{niece}(a, c) ~\leftarrow~ \text{brother}(a, b) \wedge \text{daughter}(b, c).
\end{align*}
Note that the structure of the above rule can be captured by a higher-order logical predicate called ``composite'' (abbreviated as \sclcode{comp}).
This allows us to express many other similarly structured rules with ease.
For instance, we can have \sclcode{comp(brother, daughter, niece)} and \sclcode{comp(father, mother, grandmother)}.
With this set of rules, we may derive more facts based on known kinship relations.
In fact, composition is the only kind of rule we need for kinship reasoning.
In general, there are many other useful higher-order predicates to reason over knowledge bases, which we list out in Table \ref{tab:higher-order-predicate}.

\begin{table}[h]
    \centering
    \footnotesize
    \begin{tabular}{c|c}
        \toprule \hline
        \textbf{Predicate} & \textbf{Example} \\ \hline
        transitive & $\text{transitive}(\text{relative})$ \\ \hline
        symmetric & $\text{symmetric}(\text{spouse})$ \\ \hline
        inverse & $\text{inverse}(\text{husband}, \text{wife})$ \\ \hline
        implies & $\text{implies}(\text{mother}, \text{parent})$ \\
        \bottomrule
    \end{tabular}
    \caption{Higher-order predicate examples.}
    \label{tab:higher-order-predicate}
    \vspace{-15px}
\end{table}

\paragraph{Probability propagation.}

We seek to have the deduced facts to also be associated with probabilities computed using probabilities predicted by the underlying relation extractor $\mathcal{M}_\theta$.
This is achieved by allowing the propagation of probabilities.
For example, we have the proof tree with probabilities:
\begin{align*}
\small
\inferrule
    {0.9::\text{brother}(D, R) \\ 0.8::\text{daughter}(R, K)}
    {0.72::\text{niece}(D, K)}
\end{align*}
In practice, there could be multiple steps in the proof tree (multi-hop) and one fact can be derived by multiple proof trees.
We employ the inference algorithms based on approximated \textit{weighted model counting} (WMC) presented in \cite{manhaeve2018deepproblog} to account for probabilistic inference under complex scenarios.
Since the WMC procedure is augmented for differentiation, we can obtain the gradient $\frac{\partial \hat{y}}{\partial \mathbf{r}}$.
From here, we can obtain $\frac{\partial \hat{y}}{\partial \theta} = \frac{\partial \hat{y}}{\partial \mathbf{r}} \frac{\partial \mathbf{r}}{\partial \theta}$, where the second part can be automatically derived from differentiating $\mathcal{M}_\theta$.

\paragraph{Rule learning.}
Hand-crafted rules could be expensive or even impossible to obtain.
To alleviate this issue, \abbr~applies LMs to help automatically extract rules, and further utilizes the differentiable pipeline to fine-tune the rules.
Each rule such as \sclcode{comp(brother, daughter, niece)} is attached a weight, initialized by prompting an underlying LM.
For example, the prompt we use for extracting \sclcode{comp($r$,$p$,$q$)} is ``one's $r$'s $p$ is their \texttt{<}$q$\texttt{:mask>}''.
Given that the relations $r, p, q \in \R$, \abbr~ automatically enumerates $r$ and $p$ from $\R$ while querying for LM to unmask the value of $q$.
LM then returns a distribution of words, which we take an intersection with $\R$.
The probabilities combined form the initial rule weights $\phi$.
This type of rule extraction strategy is different from existing approaches in inductive logic programming since we are exploiting LMs for existing knowledge about relationships.

Note that LMs often make simple mistakes answering such prompt.
In fact, with the above prompt, even GPT-3 can only produce 62\% of composition rules correctly.
While we can edit prompt to include few-shot examples, in this work we consider fine-tuning such rule weights $\phi$ within our differentiable reasoning pipeline.
The gradient with respect to $\phi$ is also derived with the WMC procedure, giving us $\frac{\partial \hat{y}}{\partial \phi}$.
In practice, we use two optimizers with different hyper-parameters to update the rule weights $\phi$ and the underlying model parameter $\theta$, in order to account for optimizing different types of weights.

\paragraph{Semantic loss and integrity constraints.}
In general, learning with weak supervision label is hard, not to mention that the deductive rules are learnt as well.
We thereby introduce an additional semantic loss during training.
Here, semantic loss is derived by a set of integrity constraints used to regularize the predicted entity-relation graph as well as the learnt logic rules.
In particular, we consider rules that detect \textit{violations} of integrity constraints.
For example, ``if A is B's father, then B should be A's son or daughter'' is an integrity constraint for relation extractor---if the model predicts a father relationship between A and B, then it should also predict a son or daughter relationship between B and A.
Encoded in first order logic, it is
\begin{align*}
\small
\forall a, b, \text{father}(a, b) \Rightarrow (\text{son}(b, a) \vee \text{daughter}(b, a)).
\end{align*}
Through differentiable reasoning, we evaluate the probability of such constraint being violated, yielding our expected \textit{semantic loss}.
In practice, arbitrary number of constraints can be included, though too many interleaving ones could hinder learning.

\vspace{-2mm}

\begin{figure*}[h]
    \centering
    \scriptsize
\begin{lstlisting}[language=scallop,frame=single]
// Relation declaration
type kinship(rela: String, subject: String, object: String)
type query(subject: String, object: String)
type composite(r1: String, r2: String, r3: String)
// Rules to derive the final answer
rel kinship(r3,a,c) = kinship(r1,a,b), kinship(r2,b,c), composite(r1,r2,r3), a != c
rel answer(r) = query(s, o), derive(r, s, o)
// Integrity constraints (6 for kinship graph and 2 for rule learning)
rel violation(!r) = r := forall(a, b: kinship(FATHER, a, b) => 
  kinship(SON, b, a) or kinship(DAUGHTER, b, a)) // Other constraints are omitted...
\end{lstlisting}
\vspace{-10px}
    \caption{The Scallop program used in the CLUTRR reasoning task.}
    \label{fig:scallop_program}
    \vspace{-10px}
\end{figure*}

\section{Experiments}

We evaluate \abbr~on both CLUTRR and DBpedia-INF.
We show that \abbr~has accurate and generalizable long-range reasoning capability.

\subsection{Datasets}
\textbf{CLUTRR} \citep{sinha2019clutrr} consists of kinship reasoning questions.
Given a context that describes a family's routine activity, the goal is to deduce the relationship between two family members that is not explicitly mentioned in the story.
Although the dataset is synthetic, the sentences are crowd-sourced and hence there is a considerable amount of naturalness inside the dataset.
The family kinship graph is synthetic and the names of the family members are randomized.
For ablation study, we manually crafted 92 kinship composition rules as an external symbolic knowledge base.
This yields the following symbolic information for each datapoint:
1) the full kinship graph corresponding to the story,
2) the symbolic knowledge base (KB), and
3) a query representing the question.
The CLUTRR dataset is divided into different difficulties measured by $k$, the number of facts used in the reasoning chain.
For training, we only have 10K data points with 5K $k = 2$ and another 5K $k = 3$, meaning that we can only receive supervision on data with short reasoning chains.
The test set, on the other hand, contains 1.1K examples with $k \in \{2, \dots, 10\}$.

\textbf{DBpedia-INF} is a curated subset of the evaluation dataset used in RuleBert \cite{saeed2021rulebert}.
Similar to CLUTRR, it is generated synthetically to test the reasoning capability of LMs.
Given a synthetic passage describing the relation between entities, and soft deductive logic rules, we aim to deduce the relationship between any two entities.
The symbolic program of DBpedia-INF consists of 26 predicates, 161 soft rules mined from DBpedia, and 16 rules defining the negation and symmetricity between the predicates.
The difficulty of the questions is represented in terms of reasoning length from $k \in \{0, \dots, 5\}$.\footnote{A length of 0 means that the hypothesis can be verified using the facts alone without using any rules.}
Larger $k$ implies harder question.
Compared to the exact dataset used in Rulebert, we clean it in order to ensure the question-answer pairs are logically consistent and probabilistically correct.

\subsection{Experimental Setup}

\paragraph{Implementation.}
We employ Scallop \cite{huang2021scallop} as the differentiable symbolic inference module.
We show the program used for CLUTRR reasoning task in Figure \ref{fig:scallop_program}.
It comprises relation type declarations, deductive rules for kinship reasoning, and integrity constraints for computing semantic loss (attached in the Appendix).
The program used for DBpedia-INF is written in a similar manner with additional high-order predicates listed in Table \ref{tab:higher-order-predicate}.

\paragraph{Pre-trained LMs for fine-tuning.}
We used the HuggingFace \citep{wolf2019huggingface} pre-trained \textit{w2v-google-news-300}, RoBERTa-base, and DeBERTa-base as the pretrained language models.
We finetune RoBERTa-base and DeBERTa-base during training with binary cross entropy loss.
Our relation extraction module is implemented by adding an MLP classifier after the LM, accepting a concatenation of the embedding of the two entities and the embedding of the whole windowed context.

\paragraph{Our model.}
Our main model, DSR-LM, uses RoBERTa as the underlying LM.
The relation classifier is a 2-layer fully connected MLP.
For training, we initialize $\phi$ by prompting the LM.
To accelerate the learning process, we use multinomial sampling to retrieve 150 rules for symbolic reasoning.
During testing, we will instead pick the top 150 rules.
We use two Adam optimizer to update $\theta$ and $\phi$, with learning rate $10^{-5}$ and $10^{-2}$ respectively.

For ablation studies, we present a few other models.
First, we ablate on back-bone LMs.
Specifically, we have DSR-LM-DeBERTa which uses DeBERTa as the back-bone LM.
DSR-w2v-BiLSTM, on the other hand, uses as back-bone the word2vec \cite{tomas2013w2v} model for word embedding and BiLSTM \cite{huang2015bilstm} for sequential encoding.
For DSR-LM-with-Manual-Rule we treat the logic rules as given, meaning that we provide 92 composition rules for CLUTRR and around 180 rules for DBpedia-INF.
In this case, we set ground truth rules to have $1.0$ weight and therefore $\phi$ is not learnt.
Then, we have DSR-LM-without-IC which does not have integrity constraints and semantic loss.
Lastly, we have DSR-without-LM that takes ground truth structured entity relation graph as input.
This way, we do not need the underlying relation extractor and only $\phi$ needs to be learned.

\paragraph{Baselines.} We compare \abbr~with a spectrum of baselines from purely neural to logically structured. The baselines include pretrained large language models (BERT~\citep{kenton2019bert} and RoBERTa~\citep{liu2019roberta}), non-LM counterparts (BiLSTM~\citep{hochreiter1997long, cho2014learning} and BERT-LSTM), structured models (GAT~\citep{velivckovic2018graph}, RN~\citep{santoro2017simple}, and
MAC~\citep{hudson2018compositional}), and other neuro-symbolic models (CTP \cite{minervini2020learning}, RuleBert~\citep{saeed2021rulebert}). The structured models include those models with relational inductive biases, while the neuro-symbolic model uses logic constraints.

\paragraph{Baseline setup.}
We highlight a few baselines we include for completeness but are treated as unfair comparison to us: GAT, CTP, and GPT-3 variants.
All baselines other than GAT and CTP take as input natural language stories and the question to produce the corresponding answer.
GAT and CTP, on the contrary, takes entity relation graph rather than natural language during training and testing.

The model sizes are different across baselines as well.
Model size generally depends on two parts, the backbone pre-trained LM, and the classification network built upon the LM.
GPT-3 contains 175B parameters, and RoBERTa uses 123M parameters.
The classification model of our method has 2.97M parameters (assuming using embeddings from RoBERTa).
With extra 10K parameters for rule weights, our DSR-LM framework has around 127M parameters.

For GPT-3 variants, we conduct experiments on CLUTRR with GPT-3 under the Zero-Shot (GPT-3 ZS), GPT-3 Fine-Tuned (GPT-3 FT), and Few(5)-Shot (GPT-3 5S) \citep{brown2020language}, as well as Zero-Shot-CoT (GPT-3 ZS-CoT) \citep{kojima2022large} settings.
For fair comparison, we also include the ground truth kinship composition knowledge in GPT-3 zero shot (GPT-3 ZS w/ Rule), and 5 shot (GPT-3 5S w/ Rule).
We include the prompts we used and additional details in Appendix~\ref{app:imp}.
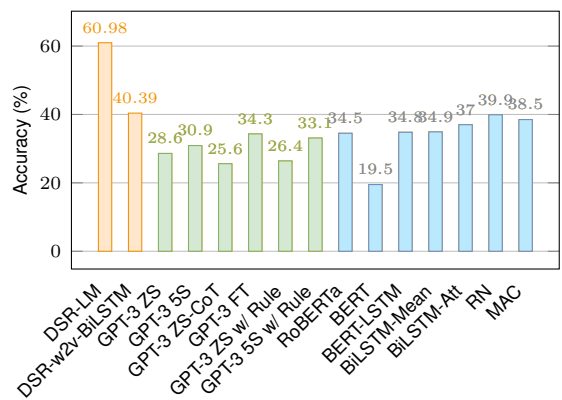
\begin{figure}[h]
    \begin{minipage}[c]{0.5\textwidth}
    \centering

\begin{tikzpicture}
\pgfplotsset{
every tick label/.append style={font=\scriptsize\sf},
}
\begin{axis}[
    height=5cm,
    width=\textwidth,
	major x tick style = transparent,
	symbolic x coords={\abbr, DSR-w2v-BiLSTM, GPT-3 ZS, GPT-3 5S, GPT-3 ZS-CoT, GPT-3 FT, GPT-3 ZS w/ Rule, GPT-3 5S w/ Rule, RoBERTa, BERT, BERT-LSTM, BiLSTM-Mean, BiLSTM-Att, RN, MAC, GAT},
    x tick style={draw=none},
	ylabel=\scriptsize{Accuracy (\%)},
	ymajorgrids = true,
	legend cell align=left,
	enlargelimits=0.08,
	xtick = data,
	ybar=\pgflinewidth/6,
	bar width=5,
    bar shift=0pt,
    nodes near coords,
	y label style={at={(axis description cs:0.09,.5)}, font={\scriptsize\sf}},
    x tick label style={rotate=45,anchor=east},
    every node near coord/.append style={font=\tiny\sf},
    ymin=0,
    ymax=65,
]

\addplot[mydeeporange, fill=myorange]
coordinates{
    (\abbr, 60.98)
    (DSR-w2v-BiLSTM, 40.39)
    (GPT-3 ZS, 28.6)
    (GPT-3 5S, 30.9)
    (GPT-3 ZS-CoT, 25.6)
    (GPT-3 FT, 34.3)
    (GPT-3 ZS w/ Rule, 26.4)
    (GPT-3 5S w/ Rule, 33.1)
    (RoBERTa, 34.5)
    (BERT, 19.5)
    (BERT-LSTM,	34.8)
    (MAC, 38.5)
    (RN, 39.9)
    (BiLSTM-Mean, 34.9)
    (BiLSTM-Att, 37.0)
};

\addplot[mydeepgreen, fill=mygreen]
coordinates{
    (GPT-3 ZS, 28.6)
    (GPT-3 5S, 30.9)
    (GPT-3 ZS-CoT, 25.6)
    (GPT-3 FT, 34.3)
    (GPT-3 ZS w/ Rule, 26.4)
    (GPT-3 5S w/ Rule, 33.1)
};

\addplot[mydeepblue, fill=myblue]
coordinates{
    (RoBERTa, 34.5)
    (BERT, 19.5)
    (BERT-LSTM,	34.8)
    (MAC, 38.5)
    (RN, 39.9)
    (BiLSTM-Mean, 34.9)
    (BiLSTM-Att, 37.0)
};

\addplot[mydeepgrey, fill=mygrey]
coordinates{
};

\end{axis}
\end{tikzpicture}
\captionof{figure}{\abbr's performance on CLUTRR compared with various baselines}
\label{fig:clutrr_overall}
\vspace{-10pt}
    \end{minipage}
\end{figure}

\begin{figure*}[t]
    \begin{minipage}[c]{0.38\textwidth}
        \centering
        \footnotesize
        \begin{tabular}{c||c|c}
    \toprule \hline
    Test Length & \abbr & RuleBert \\
    \hline
    Overall & \textbf{95.87} & 72.59 \\ \hline
    0 & \textbf{100.0} & 98.40 \\
    1 & \textbf{100.0} & 54.80 \\
    2 & \textbf{98.4} & 75.20 \\
    3 & \textbf{89.2} & 64.00 \\
    4 & \textbf{88.1} & 69.89 \\
    5 & \textbf{100.0} & 72.29 \\
    \bottomrule
\end{tabular}
\captionof{table}{DBpedia-INF generalization evaluation under different \textbf{test reasoning length}. Models are trained on 10K reasoning length $k=0$ sequences, and tested on sequences of reasoning length $k=[0, 5]$. }
\label{tab:dbpedia_inf_gen}
    \end{minipage}
    \hfill
    \begin{minipage}[c]{0.6\textwidth}
        \centering
        \footnotesize
        \begin{tabular}{c|c}
\toprule \hline
     \textbf{Confidence} & \textbf{Learnt Rules} \\ \hline

1.154  &  mother($a$,$c$) $\leftarrow$ sister($a$,$b$) $\wedge$ mother($b$,$c$) \\ \hline
1.152  &  daughter($a$,$c$) $\leftarrow$ daughter($a$,$b$) $\wedge$ sister($b$,$c$) \\ \hline
1.125  &  sister($a$,$c$) $\leftarrow$ daughter($a$,$b$) $\wedge$ aunt($b$,$c$) \\ \hline
1.125  &  father($a$,$c$) $\leftarrow$ brother($a$,$b$) $\wedge$ father($b$,$c$) \\ \hline
1.123  &  granddaughter($a$,$c$) $\leftarrow$ grandson($a$,$b$) $\wedge$ sister($b$,$c$) \\ \hline
1.120  &  brother($a$,$c$) $\leftarrow$ sister($a$,$b$) $\wedge$ brother($b$,$c$) \\ \hline
1.117  &  brother($a$,$c$) $\leftarrow$ son($a$,$b$) $\wedge$ uncle($b$,$c$) \\ \hline
1.105  &  brother($a$,$c$) $\leftarrow$ daughter($a$,$b$) $\wedge$ uncle($b$,$c$) \\ \hline
1.104  &  daughter($a$,$c$) $\leftarrow$ wife($a$,$b$) $\wedge$ daughter($b$,$c$) \\ \hline
1.102  &  mother($a$,$c$) $\leftarrow$ brother($a$,$b$) $\wedge$ mother($b$,$c$)  \\ \hline
$\dots$  &  $\dots$  \\ \bottomrule %
\end{tabular}
\captionof{table}{The learnt top-10 confident logic rules over CLUTRR.}
\label{tab:learnt_rules_top10}

    \end{minipage}
\end{figure*}

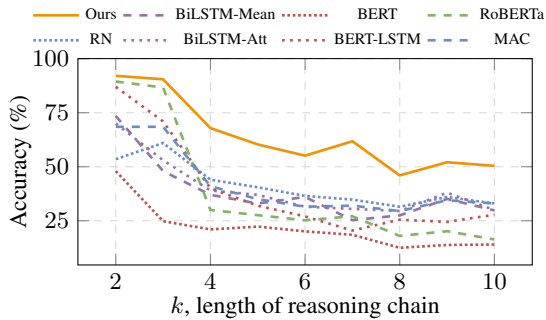
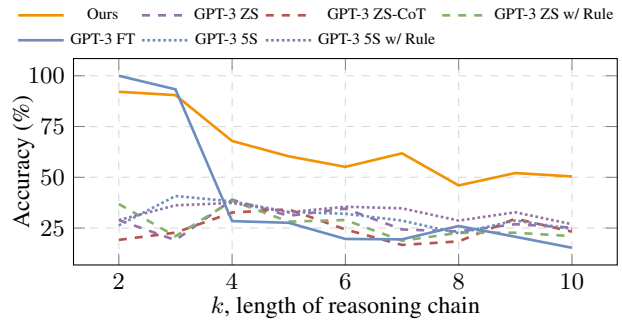
\begin{figure*}
\centering
\footnotesize
\begin{subfigure}[t]{0.45\textwidth}
\centering
\begin{tikzpicture}
      \begin{axis}[
        width=1.05\linewidth, %
        height=1.70in,
        grid=major, %
        grid style={dashed,gray!30}, %
        xlabel={$k$, length of reasoning chain},
        ylabel=Accuracy (\%),
        ytick={25,50,75,100},
        legend style={at={(-0.12,1)},draw=none,anchor=south west, font=\tiny, fill=none}, %
        x label style={at={(axis description cs:0.5,0.08)},anchor=north},
        y label style={at={(axis description cs:0.08,.5)}},
        no markers,
        every axis plot/.append style={line width=1pt},
        legend columns=4,
      ]

      \addlegendentry{Ours};
      \addplot[mydeeporange] table[x=k, y=ours, col sep=comma, ] {data/clutrr-sys-generalizability.csv};
      \addlegendentry{BiLSTM-Mean};
      \addplot[mydeeppurple, dashed] table[x=k, y=bilstm-mean, col sep=comma, ] {data/clutrr-sys-generalizability.csv};
      \addlegendentry{BERT};
      \addplot[mydeepred, densely dotted] table[x=k, y=bert, col sep=comma] {data/clutrr-sys-generalizability.csv};
      \addlegendentry{RoBERTa};
      \addplot[mydeepgreen, dashed] table[x=k, y=roberta, col sep=comma, ] {data/clutrr-sys-generalizability.csv};
      \addlegendentry{RN};
      \addplot[mydeepblue, densely dotted] table[x=k, y=rn, col sep=comma] {data/clutrr-sys-generalizability.csv};
      \addlegendentry{BiLSTM-Att};
      \addplot[mydeeppurple, dotted] table[x=k, y=bilstm-att, col sep=comma, ] {data/clutrr-sys-generalizability.csv};
      \addlegendentry{BERT-LSTM};
      \addplot[mydeepred, dotted] table[x=k, y=bert-lstm, col sep=comma] {data/clutrr-sys-generalizability.csv};
      \addlegendentry{MAC};
      \addplot[mydeepblue, dashed] table[x=k, y=mac, col sep=comma] {data/clutrr-sys-generalizability.csv};

      \end{axis}
    \end{tikzpicture}
    \caption{Comparison to common baselines}
    \label{fig:baselines}
\end{subfigure}
\hfill
\begin{subfigure}[t]{0.52\textwidth}
\centering
\begin{tikzpicture}
      \begin{axis}[
        width=1.05\linewidth, %
        height=1.70in,
        grid=major, %
        grid style={dashed,gray!30}, %
        xlabel={$k$, length of reasoning chain},
        ylabel=Accuracy (\%),
        ytick={25,50,75,100},
        legend style={at={(-0.12,1)},draw=none,anchor=south west, font=\tiny}, %
        x label style={at={(axis description cs:0.5,0.08)},anchor=north},
        y label style={at={(axis description cs:0.08,.5)}},
        no markers,
        every axis plot/.append style={line width=1pt},
        legend columns=4,
      ]

      \addlegendentry{Ours};
      \addplot[mydeeporange] table[x=k, y=ours, col sep=comma, ] {data/clutrr-sys-generalizability.csv};
      \addlegendentry{GPT-3 ZS};
      \addplot[mydeeppurple, dashed] table[x=k, y=gpt3zs, col sep=comma, ] {data/clutrr-sys-generalizability.csv};
      \addlegendentry{GPT-3 ZS-CoT};
      \addplot[mydeepred, dashed] table[x=k, y=gpt3zs-cot, col sep=comma] {data/clutrr-sys-generalizability.csv};
      \addlegendentry{GPT-3 ZS w/ Rule};
      \addplot[mydeepgreen, dashed] table[x=k, y=gpt3zs-wt, col sep=comma, ] {data/clutrr-sys-generalizability.csv};
      \addlegendentry{GPT-3 FT};
      \addplot[mydeepblue] table[x=k, y=gpt3ft, col sep=comma] {data/clutrr-sys-generalizability.csv};
      \addlegendentry{GPT-3 5S};
      \addplot[mydeepblue, densely dotted] table[x=k, y=gpt3fs, col sep=comma] {data/clutrr-sys-generalizability.csv};
      \addlegendentry{GPT-3 5S w/ Rule};
      \addplot[mydeeppurple, densely dotted] table[x=k, y=gpt3fs-wt, col sep=comma, ] {data/clutrr-sys-generalizability.csv};

      \end{axis}
    \end{tikzpicture}
    \caption{Comparison to GPT-3 related baselines}
    \label{fig:gpt3}
\end{subfigure}
\caption{\textbf{Systematic generalization} performance comparison on CLUTRR dataset.
Models except GPT-3-ZS*, GPT-3-FS are trained (or fine-tuned) on $k \in \{2, 3\}$. All models are tested on $k\in\{2,\dots,10\}$.
}
\label{fig:ablation_combo}
\vspace{-10px}
\end{figure*}

\subsection{Experimental Results}
\paragraph{\abbr~ systematically outperforms a wide range of baselines by a large margin.} %
We evaluate \abbr~and baselines on both CLUTRR and DBpedia-INF, as reported in \figref{fig:clutrr_overall} and \tabref{tab:dbpedia_inf_gen}.

In the CLUTRR experiment, \abbr~achieves the best performance among all the models (\figref{fig:clutrr_overall}).
Next, we examine how models trained on stories generated from clauses of length $k\le3$ and evaluated on stories generated from larger clauses of
length $k \geq 4$.
A fine-grained generalizability study reveals that although all models' performances decline as the reasoning length of the test sequence increases, pure neural-based models decrease the fastest (\figref{fig:baselines} and \ref{fig:gpt3}).
It manifests the systematic issue that language models alone are still not robust for length generalization \citep{lake2018generalization}.
On the other hand, the performance of \abbr~decreases much slower as test reasoning length increases and outperforms all the baselines when $k \geq 4$.

In the DBpedia-INF experiment,
\abbr~outperforms RuleBert by 37\% in terms of overall performance (\tabref{tab:dbpedia_inf_gen}), showing that \abbr~has much more robust generalization.
Recall that RuleBert aims to improve the logical reasoning of LMs by straightforward fine-tuning with soft rules and facts.
Our results show that augmenting data alone for fine-tuning do not effectively improve systematicity.
Meanwhile, \abbr~ imbues reasoning inductive biases throughout training and learns useful rules to generalize to longer reasoning lengths.

\vspace{-1mm}
\paragraph{Learning interpretable logic rules.}
\abbr~ is capable of producing explicit logic rules as part of the learning process.
For presentation, we show the top-10 rules learnt from DSR-LM model in Table \ref{tab:learnt_rules_top10}.
We compare the top-92 most likely prompted and fine-tuned rules against the 92 hand-crafted rules, and 70 of them match.
Additionally, we find that our rule weight fine-tuning helps correct 11 of the incorrect rules produced by LM.
Through this qualitative analysis, it is clear that \abbr~provides an interface to probe and interpret the intermediate steps, enhancing the interpretability.

\vspace{-1mm}
\paragraph{GPT-3 variants are inferior in long-range reasoning.}
Interestingly, ZS scores 28.6\% accuracy on CLUTRR while ZS-CoT scores 25.6\%, suggesting that the chain-of-thought prompting might not work in long-range reasoning (\figref{fig:clutrr_overall}).
In fact, there are many cases where GPT-3 favors complication over simplicity: GPT-3 frequently answers ``stepdaughter'', ``stepmother'', and ``adopted son'', while the real answers are simply ``daughter'', ``mother'', and ``son''.
Additionally, GPT-3 could derive the correct result for the wrong reason, e.g. ``Jeffrey is Gabrielle's son, which would make William her grandson, and Jeffrey's brother.''
While we count the final answer to be correct (William is Jeffrey's brother), there is a clear inconsistency in the reasoning chain:
William cannot be Gabrielle's grandson and Jeffrey's brother simultaneously, given that Jeffrey is Gabrielle's son.
Lastly, we observe that, both GPT-3 FT and many other methods have an accuracy drop as $k$ becomes larger (Figure \ref{fig:gpt3}), ZS and ZS-CoT stay relatively consistent,
suggesting that the size of context and the reasoning chain may have a low impact on GPT-3's performance.
\vspace{-1mm}
\subsection{Analyses and Ablation Studies}

\paragraph{Symbolic reasoner consistently improves LMs and word embeddings.} Since \abbr~has a model agnostic architecture, we study how the choice of different LMs impacts the reasoning performance.
As shown in \tabref{tab:clutrr_abl_lm}, the two transformer-based models have on-par performance and outperform the word2vec one.
However, note that the word2vec-based model still has better performance than all other baselines.
Besides higher final accuracy, the pre-trained transformer-based language model also accelerates the training process.
Both \abbr-RoBERTa and \abbr-DeBERTa reach their best performance within 20 epochs, while it takes DSR-w2v-BiLSTM 40 epochs to peak.

\begin{table}[h]
    \centering
    \footnotesize
    \begin{tabular}{c|c}
        \toprule \hline
        Model & Accuracy (\%) \\ \hline
        \abbr~(RoBERTa) & \textbf{60.98 $\pm$ 2.64} \\ \hline
        \abbr-DeBERTa & 60.92 $\pm$ 2.72 \\ \hline
        DSR-w2v-BiLSTM & 40.39 $\pm$ 0.06 \\
        \bottomrule
    \end{tabular}
    \caption{Ablation study about \textbf{neural backbones} of \abbr. We compare the CLUTRR performance of \abbr~using different LMs. }
    \label{tab:clutrr_abl_lm}
\end{table}

\vspace{-10px}

\paragraph{Incorporate domain knowledge.}
\abbr~allows injecting domain specific knowledge.
In \abbr-with-Rule, we manually crafted 92 rules for kinship reasoning to replace the learnt rules.
As shown in \tabref{tab:clutrr_abl_rule}, it obtained a 0.36\% performance gain over \abbr.
The fact that the improvement is marginal implies our method extracts useful rules to obtain on-par performance with manually crafted ones.
\abbr-without-IC, our model without integrity constraints specified on predicted relations and rules, performs worse than \abbr, suggesting that logical integrity constraints are essential component for improving the model robustness.

\begin{table}[h]
    \centering
    \footnotesize
    \begin{tabular}{c|c}
        \toprule \hline
        Model & Accuracy (\%) \\
        \hline
        \abbr & 60.98 $\pm$ 2.64 \\ \hline
        \abbr-without-IC & 51.48 $\pm$ 0.57 \\ \hline
        \abbr-with-Manual-Rule & \textbf{61.34 $\pm$ 1.56} \\
        \bottomrule
    \end{tabular}
    \caption{
    Ablation study. We compare our model's performance on CLUTRR with different setups.
    }
    \label{tab:clutrr_abl_rule}
    \vspace{-10px}
\end{table}

\paragraph{The impact of the relation extractor.} To understand what causes the failure case of \abbr, we study the performance of our relation classification model separately.
We isolate the trained relation extractor and found that it reaches 84.69\% accuracy on the single relation classification task.
For comparison, we train a relation extractor using all the intermediate labels in the training dataset, and it reaches 85.32\% accuracy.
It shows that even using only weak supervision (i.e., the final answers to multi-hop questions), our approach can reach on-par performance as supervised relation extraction.

\paragraph{Reasoning over structured KBs.}
To understand the rule learning capability of our approach, we design our ablation model DSR-without-LM to take as input ground-truth KBs instead of natural language.
In this case, rule weights are \textit{not} initialized by LM but randomized.
As shown in Table \ref{tab:with-knowledge-graph}, our model outperforms GAT and CTP which also operates on structured KBs.
It demonstrates that our differentiable rule learning paradigm learns rules to reason about KBs consistently.

\begin{table}[h]
    \centering
    \footnotesize
    \begin{tabular}{c|c}
        \toprule \hline
        Model & Accuracy (\%) \\ \hline
        GAT & 39.05 \\ \hline
        CTP & 95.57 \\ \hline
        DSR-without-LM & \textbf{98.81} \\
        \bottomrule
    \end{tabular}
    \caption{
        DSR-without-LM compared against GAT and CTP on reasoning with ground truth KBs.
        For this comparison we train on $k \in [2, 3]$ and test on $k \in [4, 10]$.
    }
    \label{tab:with-knowledge-graph}
    \vspace{-10pt}
\end{table}

\paragraph{Failure cases of \abbr.}
We showcase in Appendix Table~\ref{tab:fail} that even state-of-the-art large LMs are prone to logical fallacies.
On the other hand, the failure case of our method usually occurs in the stage of relation extraction.
For example, for the following sentence ``Christopher and Guillermina are having a father-daughter dance'',
our RoBERTa based relation extractor fails to recognize the father-daughter relationship but rather thinks C and G have a husband-wife relationship.
We require most of the relation extraction to be correct in order to avoid cascading error.
As the error rate on individual relation extraction accumulates, it leads to the observed drop in accuracy as $k$ becomes larger.

\vspace{-2mm}

\section{Concluding Remarks}
We investigate how to improve LMs' logical reasoning capability using differentiable symbolic reasoning. 
Through extensive experiments, we demonstrate the effectiveness of \abbr~over challenging scenarios where widely deployed large LMs fail to reason reliably.
We hope our work can lay the groundwork for exploring neuro-symbolic programming techniques to improve the robustness of LMs on reasoning problems.

\bibliography{ref}
\bibliographystyle{acl_natbib}

\appendix

\clearpage 
\newpage
\vspace{-3mm}
\section{Implementation Details}
\label{app:imp}
\vspace{-2mm}

\textbf{Hardware.}
We perform all the experiments on a server with two 20-core Intel Xeon CPUs, four GeForce RTX 2080 Ti GPUs, and 768 GB RAM.

\textbf{Reasoner details.} 
The learning of rules and the fine-tuning of the underlying LM should happen separately with different learning rates
-- fine-tuning LM is an intricate process that requires a very small learning rate, while rules should be learned with larger learning rates since gradients are directly back-propagated onto the weights.
This can be realized by employing two separate optimizers, one for fine-tuning and the other for rule learning.
During training time, we rotate training the two parts by toggling one and the other optimizer for every 10 batches of data points. 

\textbf{Rule learning training setup.}
For rule learning, we can initialize the transitivity tensor using the language model provided composite rules. 
Since the CLUTRR dataset consists of $20$ different relations and a transitivity relationship is defined over $3$ relations, there are 8K possible transitivity facts over these relations. 
Specifically, we give every predicted composite rule by the GPT with a $0.5$ weight, while initializing the other rules with a range such as $[0, 0.1]$, since otherwise, an insensible transitive fact may be getting a random high weight while it effectively does nothing for reasoning.
The learning process encourages the rules that yield the correct query result and suppresses the rules that lead to wrong answers.
To avoid the exponential blow-up caused by injecting all the 8K rules in the reasoning engine, we sample $200$ rules according to their weights during the training time and deterministically use the top $200$ learned rules during the test time.
For the \textit{QA-No-Rule} setup, the confidence score of rules, the MLP classifier for relation extraction, and the underlying LM are learned and updated simultaneously during training.
To account for their difference, we employ two Adam optimizers $A_{\text{RL}}$ and $A_{\text{RE}}$.
$A_{\text{RE}}$ is used for optimizing models for relation extraction, and thus will take as parameters the MLP classifier and the underlying LM.
It has a low learning rate $0.00001$ since it needs to fine-tune LMs.
$A_{\text{RL}}$, on the other hand, will take as a parameter the confidence score tensor for the transitive rules, and is set to have a higher learning rate of $0.001$.
For the integrity constraints, we set the result integrity violation loss with the weight $0.1$, and set the rule integrity constraint violation loss with the weight $0.01$.
We set the batch size to $16$ and train for $20$ epochs. 

To obtain the initial rule weights for the composition rule in our CLUTRR experiment, the prompt we use is ``Mary's \texttt{P}'s \texttt{Q} is her \texttt{<mask>}.'' where \texttt{P} and \texttt{Q} are enumerations of all possible relationships, and the unmasked value is treated as the answer \texttt{R}, producing \texttt{composite(P, Q, R)}.
For the other rule templates we used, the prompts are
\begin{enumerate}
\vspace{-2mm}
\item transitive: ``is \texttt{R}'s \texttt{R} one's \texttt{R}? \texttt{<mask>}''; the probability of the unmasked word being ``yes'' is treated the rule weight for 
\texttt{transitive(R)}.
\vspace{-2mm}

\item symmetric: ``does A is \texttt{R} of B means B is \texttt{R} of A? \texttt{<mask>}''; the probability of the unmasked word being ``yes'' is treated the rule weight for \texttt{symmetric(R)}.
\vspace{-2mm}
\item inverse: ``A is \texttt{R} of B means B is \texttt{<mask>} of A''; the unmasked value is treated as the answer \texttt{P}, producing \texttt{inverse(R, P)}.
\vspace{-2mm}
\item implies: ``does \texttt{R} imply \texttt{P}? <mask>''; the probability of unmasked value being ``yes'' is treated as the rule weight for \texttt{implies(R, P)}.
\end{enumerate} 

\textbf{GPT-3 Prompt Setups.}
For Zero-Shot, we use the prompt ``So $B$ is $A$'s:'' for the query pair $(A, B)$ to ask GPT-3 to complete the relationship between $A$ and $B$.
We pick the phrase in the first line or before the first period from the completed text and compare it directly with the ground truth relation.
For the Few(5)-Shot setting, we randomly select 5 examples from the training dataset used for other models ($k \in [2, 3]$) to serve as examples.
We use the same prompt for Few-Shot and Fine-Tuned as the Zero-Shot and the automated GPT-3 fine-tuning setup for our training dataset, trained for 4 epochs.
To add in the transitive KB, we simply include 92 hand-crafted rules in natural language as a part of the prompt, and we performed Zero-shot with KB, and Few(5)-shot with KB experiments.
For the Zero-Shot-CoT setting, we use the prompt ``Who is $B$ to $A$? Let's think step by step'' to suggest GPT-3 to auto-complete while working out a reasoning chain.
Under this setup, it is impossible to compare the answer to the ground truth automatically.
Therefore, we manually check through the whole test dataset of CLUTRR.

\section{Additional Experimental Results}
\label{app:exp}
In Table~\ref{tab:fail}, we showcase the failure cases of large LMs for logical inference, where Zero-shot-CoT denotes zero-shot chain-of-thoughts~\citep{zero2022kojima}.
\begin{table*}[ht]
\centering
\begin{adjustbox}{width=\textwidth,center}
\begin{tabular}{c|p{4in}|p{2in}|c}
    \toprule \hline
    $k$ & \textbf{Prompt} & \textbf{GPT-3 CoT Output} & \textbf{Ground Truth} \\ \hline
    2 &
    Jeffrey asked his brother William if he would come help him fix his car next weekend.
    Gabrielle was delighted to see her grandson, William, at her birthday party.
    Who is \textcolor{blue}{Jeffrey} to \textcolor{blue}{Gabrielle}?
    \textcolor{orange}{Let's do it step by step:}
    &
    \textcolor{blue}{Jeffrey} is \textcolor{blue}{Gabrielle}'s \textcolor{sclgreen}{grandson}.
    William is Jeffrey's brother.
    &
    \textcolor{sclgreen}{grandson} \\ \hline
    
    3 &
    Elliott felt lonely when his wife Marian was gone to see her mother Molly.
    Elliott and his daughter Karen took a day off school to go to the zoo.
    Who is \textcolor{blue}{Karen} to \textcolor{blue}{Molly}?
    \textcolor{orange}{Let's do it step by step:}
    &
    Karen is Elliott's daughter.
    Molly is Elliott's wife's mother.
    Therefore, \textcolor{blue}{Karen} is \textcolor{blue}{Molly}'s \textcolor{sclgreen}{granddaughter}.
    &
    \textcolor{sclgreen}{granddaughter} \\ \hline

    4 &
    Frank took his daughter Julia to the park. 
    Beverly, Emilie´s sister went with them. Kimberley was excited because today she was going to the zoo with her uncle Frank. 
    Jenny spent a great day shopping with her daughter, Kimberley. Who is \textcolor{blue}{Beverly} to \textcolor{blue}{Jenny}?
    \textcolor{orange}{Let's do it step by step:}
    &
    \textcolor{blue}{Beverly} is \textcolor{blue}{Jenny}'s \textcolor{red}{sister-in-law}.
    &
    \textcolor{sclgreen}{niece} \\ \hline
    
    5 &
    Melissa and her father, Stephen, went to the marina. Stephen's daughter, Margaret, had purchased a boat, and they were eager to see it. Annie and her aunt, Evelyn, went to the deli. They got a half a pound of corned beef, and two pounds of salami. Melissa and her sister Annie were arguing over a toy. While they were fighting, Melissa's sister Margaret grabbed the toy, and ran off to play with it by herself. Who is \textcolor{blue}{Evelyn} to \textcolor{blue}{Melissa}?
    \textcolor{orange}{Let's do it step by step:}
    &
    Melissa's father is Stephen. Stephen has a daughter, Margaret. Margaret is Annie's aunt. Therefore, \textcolor{blue}{Evelyn} is Annie's aunt, and \textcolor{blue}{Melissa}'s \textcolor{red}{great-aunt}.
    &
    \textcolor{sclgreen}{aunt} \\ \hline
    
    6 &
    James took his grandson Davis to the baseball game. Sharon and her son Brian went out to lunch together yesterday. Brian and his brother Oliver went to the park to play basketball. Davis played scrabble with his mother Sharon and his brother Oliver. Davis asked his brother Oliver if he would come help him fix his car next weekend. 
    Who is \textcolor{blue}{James} to \textcolor{blue}{Oliver}?
    \textcolor{orange}{Let's do it step by step:}
    &
    \textcolor{blue}{James} is Davis's grandfather, so he is \textcolor{blue}{Oliver}'s \textcolor{red}{great-grandfather}.
    &
    \textcolor{sclgreen}{grandfather} \\ \hline 
    
    7 &
    George rushed to the hospital to find out that his wife and already given birth to a boy and had named him David. Larry was playing checkers with his brother Craig. Larry's sister Jean was waiting to play. Judy made mixed drinks for her son Craig. Judy's daughter Jean was jealous because she was n't offered one. Jean planned a trip to the zoo for her brother, Craig. They had a great time. George was having trouble with his divorce because he only got to see his daughter Jean a few times a week. 
    Who is \textcolor{blue}{David} to \textcolor{blue}{Craig}?
    \textcolor{orange}{Let's do it step by step:}
    &
    George's wife gave birth to a boy and named him David. \textcolor{blue}{David} is \textcolor{blue}{Craig}'s \textcolor{red}{nephew}.
    &
    \textcolor{sclgreen}{brother} \\ \hline 
    
    8 &
    Anthony went to the bakery with his uncle Roger to pick up some bread for lunch. Anthony loved going to the store with his mom Mary. She always bought him snacks Charles would n't let his son Floyd go to the park by himself. Floyd's brother Anthony offered to go with him. Faye was hoping to travel out of state to visit her son Cecil. Ellie's brother, Charles, was doing his best to find the best flight. Mary and her son Anthony went out to lunch together yesterday. Cecil took her sister, Christine, out to dinner for her birthday. Who is \textcolor{blue}{Roger} to \textcolor{blue}{Christine}?
    \textcolor{orange}{Let's do it step by step:}
    &
    Cecil is Faye's son, so Faye is Christine's mother-in-law. Christine is married to Faye's son, so Christine is Faye's daughter-in-law. Charles is Ellie's brother, so Charles is Christine's brother-in-law. Anthony is Charles's son, so Anthony is Christine's nephew. Roger is Anthony's uncle, so \textcolor{blue}{Roger} is \textcolor{blue}{Christine}'s \textcolor{red}{brother-in-law}.
    &
    \textcolor{sclgreen}{uncle} \\ \hline 
    
    9 &
    Gino bought gifts for his lovely sister, Jami. Michele and William left that evening to go bowling. Ethel and her son, Gordon, went to the park, and had a wonderful time. Manuel's father, Kenneth, thought his son was a monster. Meanwhile Michele thought her son Manuel was a little angel. Gino, Kenneth's other son, agreed with his dad. William asked his brother Larry if he would come help him fix his car next weekend. Joseph was so happy for his brother, Larry, getting a promotion. Gordon and his uncle Joseph went to the movies Sunday after church and got popcorn and candy while they were there. Who is \textcolor{blue}{Jami} to \textcolor{blue}{Ethel}?
    \textcolor{orange}{Let's do it step by step:}
    &
    1. Jami is Gino's sister. 2. Ethel is Gino's mother. 3. Therefore, \textcolor{blue}{Jami} is \textcolor{blue}{Ethel}'s \textcolor{red}{daughter-in-law}.
    &
    \textcolor{sclgreen}{niece} \\ \hline 
    
    \bottomrule
\end{tabular}
\end{adjustbox}
\caption{
Qualitative analysis of GPT-3 Zero-Shot-CoT on the CLUTRR dataset.
The novelty comes from the sentence marked in \textcolor{orange}{orange}.
Queries that are of interest are marked in \textcolor{blue}{blue}.
Correct answer in the output is marked \textcolor{sclgreen}{green} and incorrect ones are marked \textcolor{red}{red}. 
}
\label{tab:fail}
\end{table*}

\begin{table*}[ht]
\centering
\adjustbox{max width=\textwidth}{%
\begin{tabular}{c|c}
    \toprule \hline
    \textbf{Confidence} & \textbf{Rule} \\ \hline
1.154  &  mother(A,B) $\leftarrow$ sister(A,C) $\wedge$ mother(C,B) \\ \hline
1.152  &  daughter(A,B) $\leftarrow$ daughter(A,C) $\wedge$ sister(C,B) \\ \hline
1.125  &  sister(A,B) $\leftarrow$ daughter(A,C) $\wedge$ aunt(C,B) \\ \hline
1.125  &  father(A,B) $\leftarrow$ brother(A,C) $\wedge$ father(C,B) \\ \hline
1.123  &  granddaughter(A,B) $\leftarrow$ grandson(A,C) $\wedge$ sister(C,B) \\ \hline
1.120  &  brother(A,B) $\leftarrow$ sister(A,C) $\wedge$ brother(C,B) \\ \hline
1.117  &  brother(A,B) $\leftarrow$ son(A,C) $\wedge$ uncle(C,B) \\ \hline
1.105  &  brother(A,B) $\leftarrow$ daughter(A,C) $\wedge$ uncle(C,B) \\ \hline
1.104  &  daughter(A,B) $\leftarrow$ wife(A,C) $\wedge$ daughter(C,B) \\ \hline
1.102  &  mother(A,B) $\leftarrow$ brother(A,C) $\wedge$ mother(C,B) \\ \hline
1.102  &  brother(A,B) $\leftarrow$ father(A,C) $\wedge$ son(C,B) \\ \hline
1.096  &  sister(A,B) $\leftarrow$ mother(A,C) $\wedge$ daughter(C,B) \\ \hline
1.071  &  sister(A,B) $\leftarrow$ father(A,C) $\wedge$ daughter(C,B) \\ \hline
1.071  &  son(A,B) $\leftarrow$ son(A,C) $\wedge$ brother(C,B) \\ \hline
1.070  &  uncle(A,B) $\leftarrow$ father(A,C) $\wedge$ brother(C,B) \\ \hline
1.066  &  daughter(A,B) $\leftarrow$ son(A,C) $\wedge$ sister(C,B) \\ \hline
1.061  &  brother(A,B) $\leftarrow$ brother(A,C) $\wedge$ brother(C,B) \\ \hline
1.056  &  grandson(A,B) $\leftarrow$ husband(A,C) $\wedge$ grandson(C,B) \\ \hline
1.055  &  sister(A,B) $\leftarrow$ son(A,C) $\wedge$ aunt(C,B) \\ \hline
1.053  &  grandmother(A,B) $\leftarrow$ sister(A,C) $\wedge$ grandmother(C,B) \\ \hline
1.050  &  granddaughter(A,B) $\leftarrow$ granddaughter(A,C) $\wedge$ sister(C,B) \\ \hline
1.050  &  grandmother(A,B) $\leftarrow$ brother(A,C) $\wedge$ grandmother(C,B) \\ \hline
1.047  &  grandson(A,B) $\leftarrow$ granddaughter(A,C) $\wedge$ brother(C,B) \\ \hline
1.046  &  grandfather(A,B) $\leftarrow$ mother(A,C) $\wedge$ father(C,B) \\ \hline
1.036  &  son(A,B) $\leftarrow$ daughter(A,C) $\wedge$ brother(C,B) \\ \hline
1.035  &  sister(A,B) $\leftarrow$ brother(A,C) $\wedge$ sister(C,B) \\ \hline
1.029  &  grandmother(A,B) $\leftarrow$ mother(A,C) $\wedge$ mother(C,B) \\ \hline
1.027  &  grandfather(A,B) $\leftarrow$ sister(A,C) $\wedge$ grandfather(C,B) \\ \hline
1.019  &  brother(A,B) $\leftarrow$ mother(A,C) $\wedge$ son(C,B) \\ \hline
1.017 &  granddaughter(A,B) $\leftarrow$ wife(A,C) $\wedge$ granddaughter(C,B) \\ \hline \bottomrule %
\end{tabular}}
\caption{Showcase of the learnt logic rules with top@30 confidence of \abbr~rule learning.} \label{tab:showcase-30}
\end{table*}

\begin{figure*}[t]
\begin{lstlisting}[language=scallop,frame=single]
// question :: (sub, obj) represents a question asking about relation 
// between `sub` and `obj`
type question(sub: String, obj: String)

// context :: (rela, sub, obj) represents there is a `rela`
// between `sub` and `obj`
type kinship(rela: usize, sub: String, obj: String)

// Composition rule :: (r1, r2, r3) represents compositing r1 and r2 yields r3
type composite(r1: usize, r2: usize, r3: usize)

// Constants used for defining relation properties
const DAUGHTER = 0, SISTER = 1, ..., MOTHER_IN_LAW = 19
const MALE = 0, FEMALE = 1

type gender(r: usize, gender_id: i32)
rel gender = {(DAUGHTER, FEMALE), (SISTER, FEMALE), ..., (MOTHER_IN_LAW, FEMALE)}

type gen(r: usize, gen_id: i32)
rel gen = {(DAUGHTER, -1), (SISTER, 0), ..., (MOTHER_IN_LAW, 1)}

// Composition
rel kinship(r3, x, z) = composite(r1, r2, r3), 
kinship(r1, x, y), kinship(r2, y, z), x != z

// Answer
rel answer(r) = question(s, o), kinship(r, s, o)

// Integrity constraints on results
rel violation(!r) = r := forall(a, b: kinship(GRANDFATHER, a, b) => 
  (kinship(GRANDSON, b, a) or kinship(GRANDDAUGHTER, b, a)))
rel violation(!r) = r := forall(a, b: kinship(GRANDMOTHER, a, b) => 
  (kinship(GRANDSON, b, a) or kinship(GRANDDAUGHTER, b, a))) 
rel violation(!r) = r := forall(a, b: kinship(FATHER, a, b) => 
  (kinship(SON, b, a) or kinship(DAUGHTER, b, a))) 
rel violation(!r) = r := forall(a, b: kinship(MOTHER, a, b) => 
  (kinship(SON, b, a) or kinship(DAUGHTER, b, a)))
rel violation(!r) = r := forall(a, b: kinship(HUSBAND, a, b) => kinship(WIFE, b, a))
rel violation(!r) = r := forall(a, b: kinship(BROTHER, a, b) => 
  (kinship(SISTER, b, a) or kinship(BROTHER, b, a)))

// Integrity constraints on rules
rel violation(!r) = r := forall(r1, r2, r3: 
  composite(r1, r2, r3) and gender(r2, g) => gender(r3, g))
rel violation(!r) = r := forall(r1, r2, r3: 
  composite(r1, r2, r3) and gen(r1, g1) and gen(r2, g2) => gen(r3, g1 + g2))
\end{lstlisting}
\caption{Full Scallop program including deductive rules and integrity constraints}
\end{figure*}

\end{document}